\documentclass[letterpaper]{article}
\usepackage{authblk}
\usepackage{aaai}
\usepackage{times}
\usepackage{helvet}
\usepackage{courier}
\usepackage{graphicx}
\usepackage{booktabs}
\usepackage{appendix}
\begin{document}
%
\title{Predicting Hyperkalemia in the ICU and \\Evaluation of Generalizability and Interpretability}
\setlength\titlebox{2.6in}


\author[1]{\small Gloria Hyunjung Kwak}
\author[2, 3, 4]{\small Christina Chen}
\author[5]{\small Lowell Ling}
\author[6]{\small Erina Ghosh}
\author[2, 3, 7]{\small Leo Anthony Celi}
\author[1, 8]{\small Pan Hui}

\affil[1]{\footnotesize The Hong Kong University of Science and Technology, Hong Kong, China}
\affil[2]{\footnotesize Massachusetts Institute of Technology, Cambridge, MA, USA}
\affil[3]{\footnotesize Beth Israel Deaconess Medical Center, Boston, MA, USA}
\affil[4]{\footnotesize Harvard Medical School, Boston, MA, USA}
\affil[5]{\footnotesize The Chinese University of Hong Kong, Hong Kong, China}
\affil[6]{\footnotesize Philips Research North America, Cambridge, MA, USA}
\affil[7]{\footnotesize Harvard T.H. Chan School of Public Health, Boston, MA, USA}
\affil[8]{\footnotesize University of Helsinki, Helsinki, Finland}

\maketitle
\begin{abstract}
\begin{quote}
Hyperkalemia is a potentially life-threatening condition that can lead to fatal arrhythmias. Early identification of high risk patients can inform clinical care to mitigate the risk. While hyperkalemia is often a complication of acute kidney injury (AKI), it also occurs in the absence of AKI. We developed predictive models to identify intensive care unit (ICU) patients at risk of developing hyperkalemia by using the Medical Information Mart for Intensive Care (MIMIC) and the eICU Collaborative Research Database (eICU-CRD). Our methodology focused on building multiple models, optimizing for interpretability through model selection, and simulating various clinical scenarios.
\par In order to determine if our models perform accurately on patients with and without AKI, we evaluated the following clinical cases: (i) predicting hyperkalemia after AKI within 14 days of ICU admission, (ii) predicting hyperkalemia within 14 days of ICU admission regardless of AKI status, and compared different lead times for (i) and (ii). Both clinical scenarios were modeled using logistic regression (LR), random forest (RF), and XGBoost. 
\par Using observations from the first day in the ICU, our models were able to predict hyperkalemia with an AUC of (i) 0.79, 0.81, 0.81 and (ii) 0.81, 0.85, 0.85 for LR, RF, and XGBoost respectively. We found that 4 out of the top 5 features were consistent across the models. AKI stage was significant in the models that included all patients with or without AKI, but not in the models which only included patients with AKI. This suggests that while AKI is important for hyperkalemia, the specific stage of AKI may not be as important. Our findings require further investigation and confirmation.  
\end{quote}
\end{abstract}

\section{Introduction} 
\noindent Hyperkalemia, or high serum potassium levels, is a rare but potentially life-threatening condition that may lead to fatal cardiac arrhythmias. Identifying patients at high risk for hyperkalemia may allow providers to adjust clinical management, such as avoiding potassium repletion and potassium-containing or potassium-sparing medications. Previous studies show that hyperkalemia in hospitalized patients range from K$>$5.3 to K$>$6.0 mEq/L, but for this paper, we chose the more restrictive potassium cutoff of 6.0 mEq/L
\cite{khanagavi2014hyperkalemia,acker1998hyperkalemia}.
Previous literature investigating hyperkalemia in hospitalized patients mostly focused on evaluating the association of clinical features with the development of hyperkalemia. The only predictive models that we are aware of are based on medication administration and electrocardiograms \cite{acker1998hyperkalemia,henz2008influence,eschmann2017developing,lin2020deep}.

\par Our goal is to present a methodology to build predictive models to identify patients at high risk of developing hyperkalemia using observations from the first day of ICU admission. Models were selected to maximize clinician interpretability: LR, RF, and XGBoost. Features were selected based on literature and verified by clinical expertise. While most patients with hyperkalemia in the ICU also have AKI, it is important to capture those with normal kidney function because their hyperkalemia is easier to miss.

\section{Methods and Materials}
\label{sec:methods}
\subsection{Databases}
Experiments were conducted on two publicly available databases of critical care patients: the Medical Information Mart for Intensive Care III, IV (MIMIC-III, MIMIC-IV) and the eICU Collaborative Research Database (eICU-CRD) \cite{johnson2016mimic,pollard2018eicu,mimiciv2020}. We incorporated ICU admissions: 61,532 admissions from MIMIC-III (2001 to 2012), 25,769 admissions from MIMIC-IV (2014 to 2019), and 200,859 ICU admissions from eICU-CRD (2014 and 2015). 

\subsection{Definitions}
\subsubsection{Hyperkalemia}
We used American Heart Association’s definition of moderate hyperkalemia, K$\geq$6 mEq/L, which is more restrictive than many studies, but associated with much higher mortality \cite{vanden2010part}. We filtered out erroneous lab values, such as hemolyzed specimens, by including these constraints:
(i) only one potassium in 6 hours and the result $\geq$6,
(ii) two potassium results in 6 hours and both results $\geq$6,
(iii) one potassium level $\geq$6 with calcium gluconate administration. 

\subsubsection{AKI}
AKI stage was calculated each time a creatinine or urine output was measured, according to KDIGO guidelines \cite{khwaja2012kdigo}. Baseline creatinine is defined as the lowest creatinine within the past 7 days. We used this definition because eICU does not include pre-admission labs and we wanted to be consistent in our definition.

\subsubsection{Scenarios}
We have two clinical scenarios as follows:
(i) Case 1: AKI within 7 days of admission to the ICU, followed by hyperkalemia within the next 7 days,
(ii) Case 2: Hyperkalemia within 14 days of admission to the ICU, with or without AKI. (See Appendix Figure \ref{fig:hyper})

\par For both clinical cases, the training set was composed of patients who developed hyperkalemia between admission day 1 to 14. For the test set, we selected patients who developed hyperkalemia between day 1 to 14. To investigate increasing lead times, we also selected subgroups of patients who developed hyperkalemia between day $n$ to 14 were selected as a test set ($n:2\dots4$).

\subsection{Cohort Selection}
We included the first ICU admission for all patients between the ages of 18 and 90.
The exclusion criteria are as follows:
(i) patients who have chronic kidney disease stage V or End-stage Renal disease based on ICD-9 codes,
(ii) patients who had end-of-life-discussions within 24 hours of ICU admission,
(iii) patients who had peritoneal dialysis patients at any time,
(iv) patients who had hemodialysis prior to admission to the ICU,
(v) patients who had potassium level $\geq$6 at ICU admission.

After exclusion criteria adjustment, the number of patients in this study was 43,798 for Case 1 and 83,565 for Case 2 with variables. We collected demographics data (gender, age), laboratory variables (creatinine kinase, glucose, lactate, pH, wbc, chloride, bilirubin, platelet, alanine transaminase, phosphate, hgb, serum potassium), AKI stage, fluid balance, IV fluid use (saline, hartmann, plasmalyte, dextrose 5\%, dextrose 10\%) and medications use (ACEi/ARB, diuretics, NSAID, beta blockers, steroids, potassium chloride, nitroglycerin, vasopressor, etc. See Appendix Table \ref{tab:listmed}). Features were selected based on literature review and clinician expert opinion. We selected laboratory values drawn within 24h of admission that were closest to admission time. Fluid balance was calculated for the first 24h. Drug usage was determined positive if it was used within a day of admission. Missing values were estimated based on data from 12h before and 48h after admission (closest to admission time) and then interpolated with k-nearest neighbor (n$=$3). 

\subsection{Model}
For each of the two clinical cases, we built three models. We started with a baseline LR, which is commonly used in medical literature and well understood by clinicians. We also used RF and XGBoost, which are good options for our sparse data based on a single time point with the added benefit of being easier to interpret. 
\par After data normalization, the entire cohort of patients was Case 1: 43,798 (hyperkalemia:1,048), Case 2: 83,565 (hyperkalemia:1,821). Random shuffling and splitting were repeated for the training (60\%) and test (40\%) sets to evaluate stability using k-fold cross validation (k$=$4). Models were trained with balancing the class frequency, and parameters (number of estimators and maximum depth of trees) were chosen based on convergence of error rates. The Area under the curve (AUC)s of the receiver operating curve (ROC) were used to assess the performance of LR, RF, and XGBoost over different lead times for the test sets across the scenarios. The importance of features in this project is interpreted with local model-agnostic SHAP (SHapley Additive exPlanation) values \cite{lundberg2017unified}. SHAP values attribute to each feature the change in the expected model prediction when conditioning on that feature. The baseline characteristics of each cohort are shown in Appendix Table \ref{tab:baseline}.  

\begin{table}[htbp]
\caption{Model performance (AUC) comparison for machine learning classifiers.; LR, Logistic Regression; RF, Random Forest; XGB, XGBoost; Testdate, Test set-up date; AUC, area under the curve.} 
\begin{tabular}{cccc}
  \toprule
  {Testdate} & {Model} & 
  {AKI Cohort}&{General Cohort}\\
  {}&{}&{(Case 1)} & {(Case 2)}\\
  \midrule
  {1st$\sim$14th}&{LR}&{0.79 (0.77$-$0.81)}&{0.81 (0.80$-$0.82)}\\
  &{RF}&{0.81 (0.80$-$0.82)}&{0.85 (0.84$-$0.85)}\\
  &{XGB}&{0.81 (0.79$-$0.82)}&{0.85 (0.85$-$0.86)}\\
  {2nd$\sim$14th}&{LR}&{0.75 (0.74$-$0.76)}&{0.72 (0.71$-$0.74)}\\
  &{RF}&{0.78 (0.77$-$0.79)}&{0.80 (0.79$-$0.81)}\\
  &{XGB}&{0.78 (0.76$-$0.79)}&{0.80 (0.78$-$0.81)}\\
  {3rd$\sim$14th}&{LR}&{0.70 (0.69$-$0.72)}&{0.71 (0.70$-$0.72)}\\
  &{RF}&{0.73 (0.72$-$0.74)}&{0.80 (0.79$-$0.81)}\\
  &{XGB}&{0.73 (0.72$-$0.74)}&{0.80 (0.78$-$0.81)}\\
  {4th$\sim$14th}&{LR}&{0.70 (0.69$-$0.71)}&{0.72 (0.71$-$0.73)}\\
  &{RF}&{0.74 (0.71$-$0.76)}&{0.80 (0.78$-$0.82)}\\
  &{XGB}&{0.73 (0.70$-$0.75)}&{0.80 (0.78$-$0.82)}\\

  \bottomrule
\end{tabular}
\label{tab:performance}
\end{table}

\begin{figure}[!htbp]
    \centering
    \includegraphics[width=0.7 \linewidth]{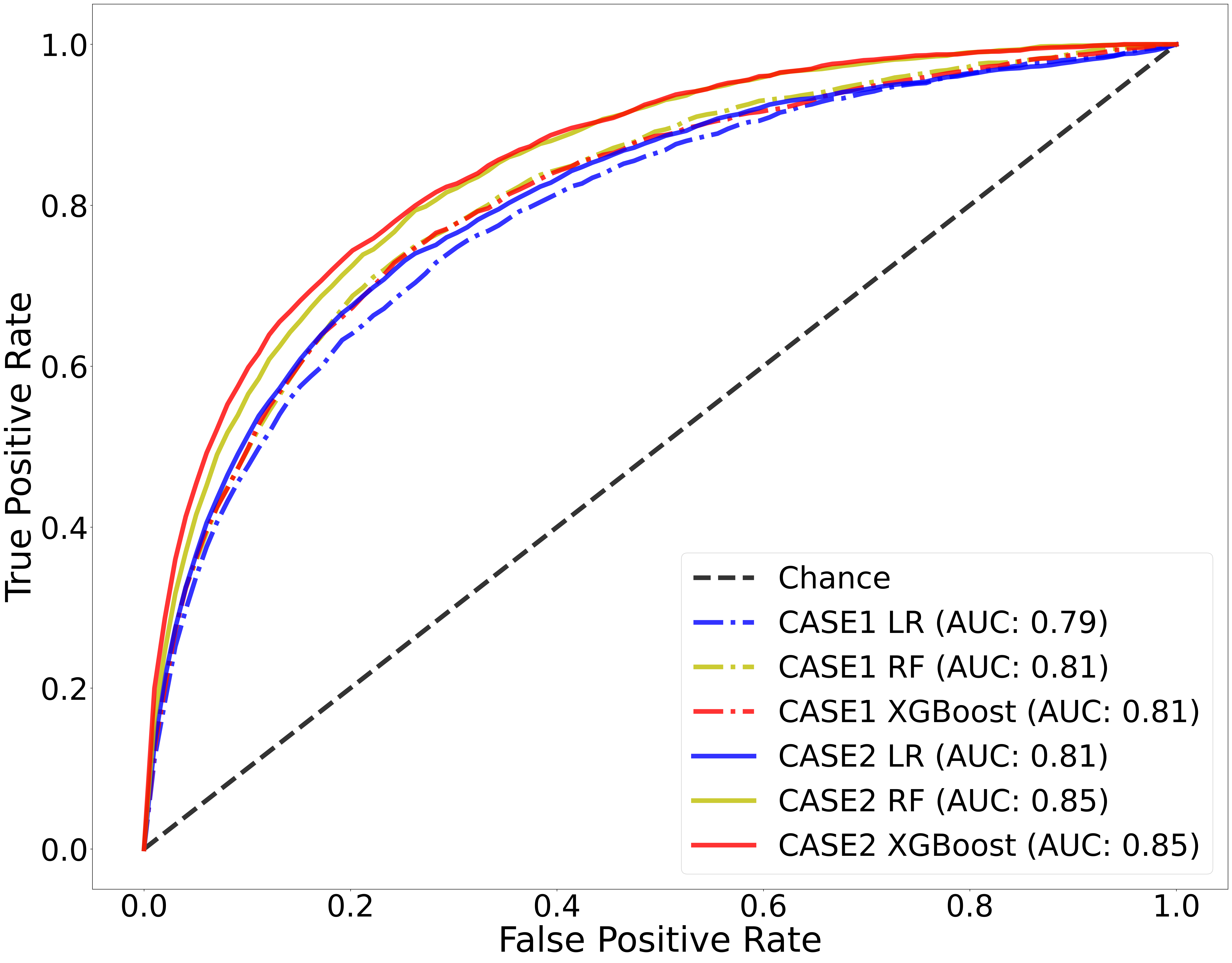}
    \caption{Model performance (AUC) comparison for machine learning classifiers.; LR, Logistic Regression; RF, Random Forest; AUC, area under the curve.}
    \label{fig:performance}
\end{figure}

\section{Results}
LR had AUC 0.79 (95\% Confidence Interval: 0.77$-$0.81) for Case 1 (AKI cohort) and 0.81 (0.80$-$0.82) for Case 2 (general cohort) with the test set-up date which is 1st to 14th day from ICU admission. RF and XGBoost performed better with AUC 0.81 (0.80$-$0.82, 0.79$-$0.82) and 0.85 (0.84$-$0.85, 0.85$-$0.86) for Case 1 and Case 2 with the same test date range (See Table \ref{tab:performance}, Figure \ref{fig:performance}). The performance of RF and XGBoost in Case 2 is consistently higher than models in Case 1. In addition, performance decreased in both cases when hyperkalemia occurred later in the hospitalization. AUC was reduced over time in both scenarios. 
\par
\begin{figure}[!htbp]
    \centering
    \includegraphics[width=0.43 \linewidth]{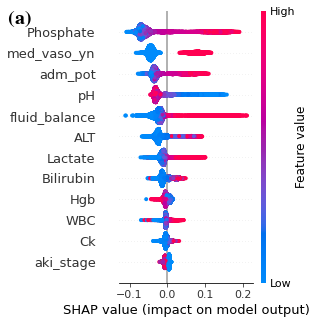} 
    \includegraphics[width=0.45 \linewidth]{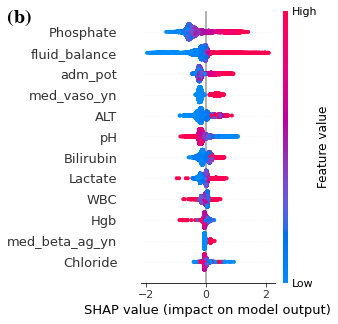} 
    \includegraphics[width=0.44 \linewidth]{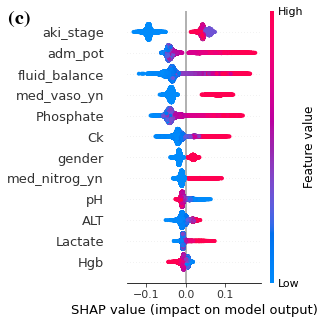} 
    \includegraphics[width=0.44 \linewidth]{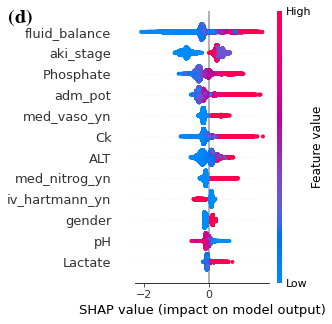} 
    \caption{Top 10 SHAP values from RF and XGBoost for Case 1 and 2. (a) RF(Case1) (b) XGBoost(Case1) (c) RF(Case2) (d) XGBoost(Case2)}
    \label{fig:SHAP}
\end{figure}

 We ran our models with and without the AKI stage as a feature, and found that all of our models were in close agreement. Analysis of feature importance is shown in Figure \ref{fig:SHAP}. Top features in RF and XGBoost models in both clinical cases include high phosphate, high admission potassium, high fluid balance, and vasopressor use. In addition, AKI stage was also an important feature in Case 2. 

\section{Discussion}
\par Using MIMIC and eICU-CRD, we built models that may be used to predict risk of hyperkalemia in critically ill patients with and without AKI. The models require parameters from the first day in the ICU to predict development of hyperkalemia within the first two weeks from admission. 
\par In both clinical scenarios, the performance of our RF and XGBoost models decreased with increasing intervals from admission time. This is likely due to the relatively longer duration of forward prediction compared to using only admission parameters for prediction, suggesting that data from subsequent, time-varying clinical states after admission day may be required to improve model prediction.
\par Top features in RF and XGBoost models in both clinical cases include high phosphate, high admission potassium, high fluid balance, and vasopressor use. Vasopressor use and positive fluid balance is suggestive of hemodynamic instability and over aggressive volume resuscitation is associated with increased morbidity and mortality. Whilst these patients had hemodynamic instability and aggressive volume resuscitation likely due to higher severity of illness, interestingly that these factors remain important even when AKI staging is featured in the model. More specifically, a positive balance is associated with hyperkalemia. This suggests treatment of shock with vasopressor and fluid therapy may alter risk of hyperkalemia beyond that of AKI as part of multiorgan dysfunction. For example, patients who require vasopressors for cardiogenic shock due to heart failure will usually be given diuretics to achieve negative fluid balance and eliminate potassium through the urine, whereas patients with hemodynamically unstable patients (ex. septic patients) often require fluid loading to improve cardiac output. In contrast, high serum phosphate and serum potassium levels on admission suggest renal dysfunction prior to admission since the kidney is largely responsible for regulating both. High phosphate is often seen in chronic kidney disease, which can impair the kidney’s ability to excrete potassium even in the absence of AKI. 
\par The AKI stage is not important in Case 1, where all patients have AKI, but it is important in Case 2. This could mean that AKI, regardless of stage, increases risk of hyperkalemia, but the difference in AKI stage does not have a large impact on risk of hyperkalemia. There are many potential reasons for this, including small cohort size and the KDIGO definition of AKI severity \cite{ostermann2020controversies}. This requires further investigation and confirmation using larger datasets.
\par Medications have been shown to have strong associations with hyperkalemia, but this was not the case for our study \cite{khanagavi2014hyperkalemia,nyirenda2009hyperkalaemia,uijtendaal2011frequency}. This could potentially be due to the severity of illness in the ICU population and the existence of other powerful causes of hyperkalemia. 

\par As emphasized in a recent commentary \cite{futoma2020myth}, a deeper understanding of the patterns discovered in clinical datasets to infer causation is necessary prior to adoption rather than simply evaluating algorithms on multiple datasets. 
In addition, an algorithm requires validation and re-calibration using local data before implementation; generalizability should never be inferred. We used multicenter datasets to create a large patient cohort of more than 83,000 patients and minimize overfitting by using broad inclusion criterias. The models might help elucidate causes of hyperkalemia in the ICU, especially those that are actionable.

\section{Conclusion}
\par We developed models to predict hyperkalemia in critically ill patients, with a focus on applicability to various clinical scenarios and interpretability. We used multi-center databases, compared multiple models optimized for interpretability, and performed sensitivity analyses using multiple use cases. 



\label{sec:cite}
\bibliography{aaai}
\bibliographystyle{aaai}

\appendix
\section{Appendix}

\begin{figure}[htbp]
  \centering{%
    \includegraphics[width=0.43\linewidth]{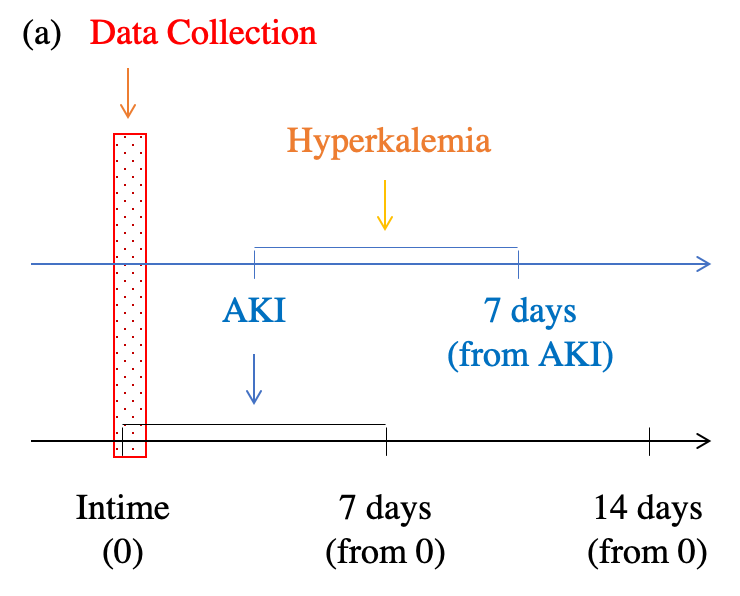}
    \includegraphics[width=0.43\linewidth]{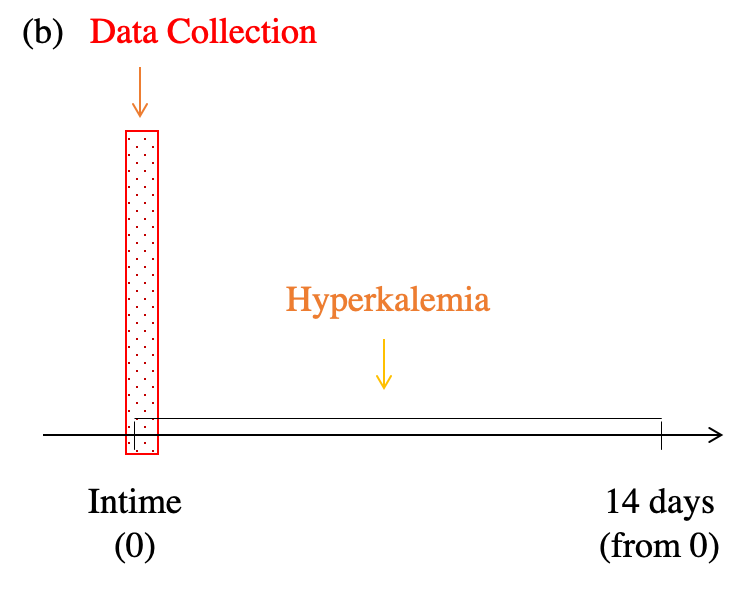}}
  \caption{Data timeframe for predicting hyperkalemia. (a) Case1 (b) Case2}
  \label{fig:hyper}
\end{figure}

\begin{table*}
\caption{List of medications}
\begin{center}
\begin{tabular}{lll}
  \toprule
  {Variable} & {Category} & {Medications}\\
  \midrule
  med$\_$ace$\_$yn & ACEi/ARB & {benazepril, monopril, captopril, enalapril, }\\
   & & {enalaprilat, lisinopril, moexipril, quinapril, }\\
   & & {ramipril, trandolapril, valsartan, }\\
   & & {losartan, irbesartan}\\
   
  med$\_$loop$\_$yn & Loop/Thiazide Diuretics & {torsemide, furosemide, chlorothiazide, }\\
   & & {indapamide, hydrochlorothiazide}\\
  med$\_$nsaid$\_$yn & NSAID & {celecoxib, celebrex, diclofenac, ibuprofen, }\\
   & & {indomethacin, ketorolac + toradol,}\\
   & & {naproxen}\\
  med$\_$beta$\_$yn & Beta Blockers & {carvedilol, esmolol, metoprolol, nadolol, }\\
   & & {propranolol, sotalol, acebutolol, atenolol, }\\
   & & {bisoprolol}\\
  med$\_$steroids$\_$yn & Steroids & {hydrocortisone na succ., methylprednisolone, }\\
   & & {prednisone}\\
  med$\_$beta$\_$ag$\_$yn & Beta Agonist & {albuterol, salmeterol, levalbuterol}\\
  med$\_$k$\_$sparing$\_$yn & K sparing Diuretics & {spironolactone, amiloride}\\
  med$\_$carbonic$\_$yn & Carbonic Anhydrase & {acetazolamide, methazolamide}\\
  & Inhibitors &\\
  med$\_$dig$\_$yn & Digoxin & \\
  med$\_$hep$\_$yn & Heparin & \\
  med$\_$pot$\_$chl$\_$yn & Potassium Chloride & \\
  med$\_$succ$\_$yn & Succinylcholine & \\
  med$\_$ins$\_$yn & Insulin & \\
  med$\_$sod$\_$bic$\_$yn & Sodium Bicarbonate & \\
  med$\_$cal$\_$yn & Calcium Gluconate & \\
  med$\_$nitrog$\_$yn & Nitroglycerin & \\
  med$\_$labet$\_$yn & Labetalol & \\
  med$\_$vasop$\_$yn & Vasopressor & {vasopressin, dopamine, phenylephrine, }\\
   & & {epinephrine, norepinephrine}\\
\bottomrule
\end{tabular}
\end{center}
\label{tab:listmed}
\end{table*}

\begin{table*}
\caption{Baseline characteristics; All values are expressed in the median and interquartile ranges unless specified; HyperK, Hyperkalemia; CKD, Chronic Kidney Disease; Comorbidities, Baseline Comorbidities; COPD, Chronic Obstructive Pulmonary Disease; SOFA, sequential organ function assessment; CK, Creatinine Kinase; Potassium, Admission potassium; Dialysis, Any form of renal replacement therapy during ICU stay; LOS, Length of Stay.}
\begin{center}
\begin{tabular}{lcccc}
  \toprule
  {} & \multicolumn{2}{c}{AKI Cohort (Case 1)} & \multicolumn{2}{c}{General Cohort (Case 2)}\\
  {} & \multicolumn{2}{c}{n=43,798} & \multicolumn{2}{c}{n=83,565}\\
  {} & {HyperK} & {Non-HyperK} &  {HyperK} & {Non-HyperK}\\
  {} & {n=1,048} & {n=42,750} &  {n=1,821} & {n=81,744}\\
  \midrule
  {Age (years)}&{64 (52.1-74.8)}&{66 (55-76.3)}&{64 (53-89)}&{64 (52-75)}\\
  {Male (\%)}&{672 (64.1\%)}&{
24,442 (57.2\%)}&{1,205 (66.2\%)}&{46,160 (56.5\%)}\\
  {SOFA score}&{8 (5-11)}&{4 (2-7)}&{6 (4-9)}&{3 (2-6)}\\
  {Comorbidities}&{}&{}&{}&{}\\
  {  CKD (\%)}&{203 (19.4)}&{4,256 (10.0)}&{297 (16.3)}&{6,074 (7.4)}\\
  {  COPD (\%)}&{96 (9.2)}&{3,945 (9.2)}&{156 (8.6)}&{6,129 (7.5)}\\
  {  Diabetes (\%)}&{172 (16.4)}&{6,307(14.8)}&{314 (17.2)}&{10,876 (13.3)}\\
  {  Hypertension (\%)}&{169 (16.1)}&{10,272 (24.0)}&{368 (20.2)}&{16,937 (20.7)}\\
  {  Stroke (\%)}&{26 (2.5)}&{1,489 (3.5)}&{43 (2.4)}&{3,012 (3.7)}\\
  {Specialty}&{}&{}&{}&{}\\
  {  Medical (\%)}&{582 (55.5)}&{23,029 (53.9)}&{784 (43.1)}&{41,518 (50.8)}\\
  {  Surgery (\%)}&{356 (34.0)}&{14,321 (33.5)}&{860 (47.2)}&{27,257 (33.3)}\\
  {  Others (\%)}&{110 (10.5)}&{5,400 (12.6)}&{177 (9.7)}&{12,969 (15.9)}\\
  {CK (U/L)}&{245 (83-1,125)}&{150 (67-447)}&{231 (84-939)}&{146 (68-434)}\\
  {Creatinine (mg/dL)}&{1 (0.7–1.5)}&{1 (0.7–1.4)}&{0.9 (0.7-1.3)}&{0.9 (0.7-1.3)}\\
  {Phosphate (mg/dL)}&{4.5 (3.4-6.1)}&{3.5 (2.8-4.2)}&{4.1 (3.2-5.3)}&{3.4 (2.7-4)}\\
  {Potassium (mEq/L)}&{4.5 (3.9-5.2)}&{4.1 (3.7-4.5)}&{4.5 (0-1)}&{4.0 (0-2)}\\
  {Calcium (mg/dL)}&{8.4 (7.8-8.9)}&{8.4 (7.8-8.9)}&{8.4 (7.9-8.9)}&{8.4 (7.9-8.9)}\\
  {Vasopressor}&{564 (53.8\%)}&{9,702 (22.7\%)}&{820 (45.0\%)}&{13,848 (16.9\%)}\\
  {Dialysis}&{433 (41.3\%)}&{2,805 (6.6\%)}&{478 (26.3\%)}&{3,088 (3.8\%)}\\
  {ICU LOS (days)}&{3 (1-9)}&{2 (1-4)}&{3 (2-7)}&{1 (1-3)}\\
  {Hospital mortality (\%)}&{326 (31.1\%)}&{3,366 (7.9\%)}&{367 (20.2\%)}&{4,532 (5.5\%)}\\
  \bottomrule
  \end{tabular}
\end{center}
\label{tab:baseline}
\end{table*}

\end{document}